\documentclass[conference,palatino]{IEEEtran}
\usepackage{algorithm}
\usepackage{amsmath}
\usepackage{booktabs}
\usepackage[dvipsnames]{xcolor}
\usepackage{colortbl}
\usepackage{textcomp}
\usepackage{xcolor}
\usepackage{hyperref}	
\usepackage{tikz}
\usepackage{comment}
\usetikzlibrary{automata, positioning, arrows}
\usepackage[singlelinecheck=false ]{caption}
\tikzset{
->, 
>=stealth,
node distance=3cm, 
every state/.style={thick, fill=gray!10}, 
initial text=$ $,
}
\usepackage{algpseudocode}

\usepackage{newfloat}
\usepackage{listings}

\lstdefinelanguage{ATNGrammar}{
  morekeywords={Start, Token, matches, Premise, Claim, Definition, State, Augmentation, NoArgument, End, does, not, match, Content, Conjunction, Action, Classify, as},
  keywordstyle=\bfseries,
  sensitive=false
}

\floatstyle{ruled}
\newfloat{listing}{tb}{lst}{}
\floatname{listing}{Listing}
\usepackage[export]{adjustbox}
\usepackage[utf8]{inputenc}
\usepackage[T1]{fontenc}
\def\BibTeX{{\rm B\kern-.05em{\sc i\kern-.025em b}\kern-.08em
    T\kern-.1667em\lower.7ex\hbox{E}\kern-.125emX}}
\usetikzlibrary{quotes}
\usepackage{array, booktabs, ragged2e}

\def\BibTeX{{\rm B\kern-.05em{\sc i\kern-.025em b}\kern-.08em
    T\kern-.1667em\lower.7ex\hbox{E}\kern-.125emX}}
\begin{document}

\title{WIBA: What Is Being Argued? A Comprehensive Approach to Argument Mining}
  
\author{\IEEEauthorblockN{Arman Irani}
\IEEEauthorblockA{\textit{UC Riverside} \\
}
\and
\IEEEauthorblockN{Ju Yeon Park}
\IEEEauthorblockA{\textit{Ohio State University} \\
}
\and
\IEEEauthorblockN{Kevin Esterling}
\IEEEauthorblockA{\textit{UC Riverside} \\
}
\and
\IEEEauthorblockN{Michalis Faloutsos}
\IEEEauthorblockA{\textit{UC Riverside}} \\
}

\maketitle

\begin{abstract}
How can we effectively model arguments communicated in diverse environments? On the one hand, there is a great opportunity with the abundance of digitized speech across different contexts including online forums, official proceedings, or transcripts of spoken debates. On the other hand, there is a great challenge in correctly detecting arguments, especially since each medium has its own set of conventions, lingo, affordances, and styles of argumentative engagement. We propose WIBA, a novel framework and suite of methods that enable the comprehensive understanding of ``\textbf{W}hat \textbf{I}s \textbf{B}eing \textbf{A}rgued" across contexts. Our approach develops a comprehensive framework that detects: (a) the existence, (b) the topic, and (c) the stance of an argument, correctly accounting for the logical dependence among the three tasks. Our algorithm leverages the fine-tuning and prompt-engineering of Large Language Models. We evaluate our approach and show that it performs well in all the three capabilities. First, we develop and release an Argument Detection model that can classify a piece of text as an argument with an $F_1$ score between 79\% and 86\% on three different benchmark datasets. Second, we release a language model that can identify the topic being argued in a sentence, be it implicit or explicit, with an average similarity score of 71\%, outperforming current naive methods by nearly 40\%. Finally, we develop a method for Argument Stance Classification, and evaluate the capability of our approach, showing it achieves a classification $F_{1}$ score between 71\% and 78\% across three diverse benchmark datasets. Our evaluation demonstrates that WIBA allows the comprehensive understanding of \textbf{W}hat \textbf{I}s \textbf{B}eing \textbf{A}rgued in large corpora across diverse contexts, which is of core interest to many applications in linguistics, communication, and social and computer science. To facilitate accessibility to the advancements outlined in this work, we release WIBA as a free open access platform (wiba.dev).

\end{abstract}
\begin{IEEEkeywords}
Argument Mining, Large Language Models, Natural Language Processing
\end{IEEEkeywords}

\section{Introduction}

\begin{figure}
    \centering
    \includegraphics[scale=0.25, right]{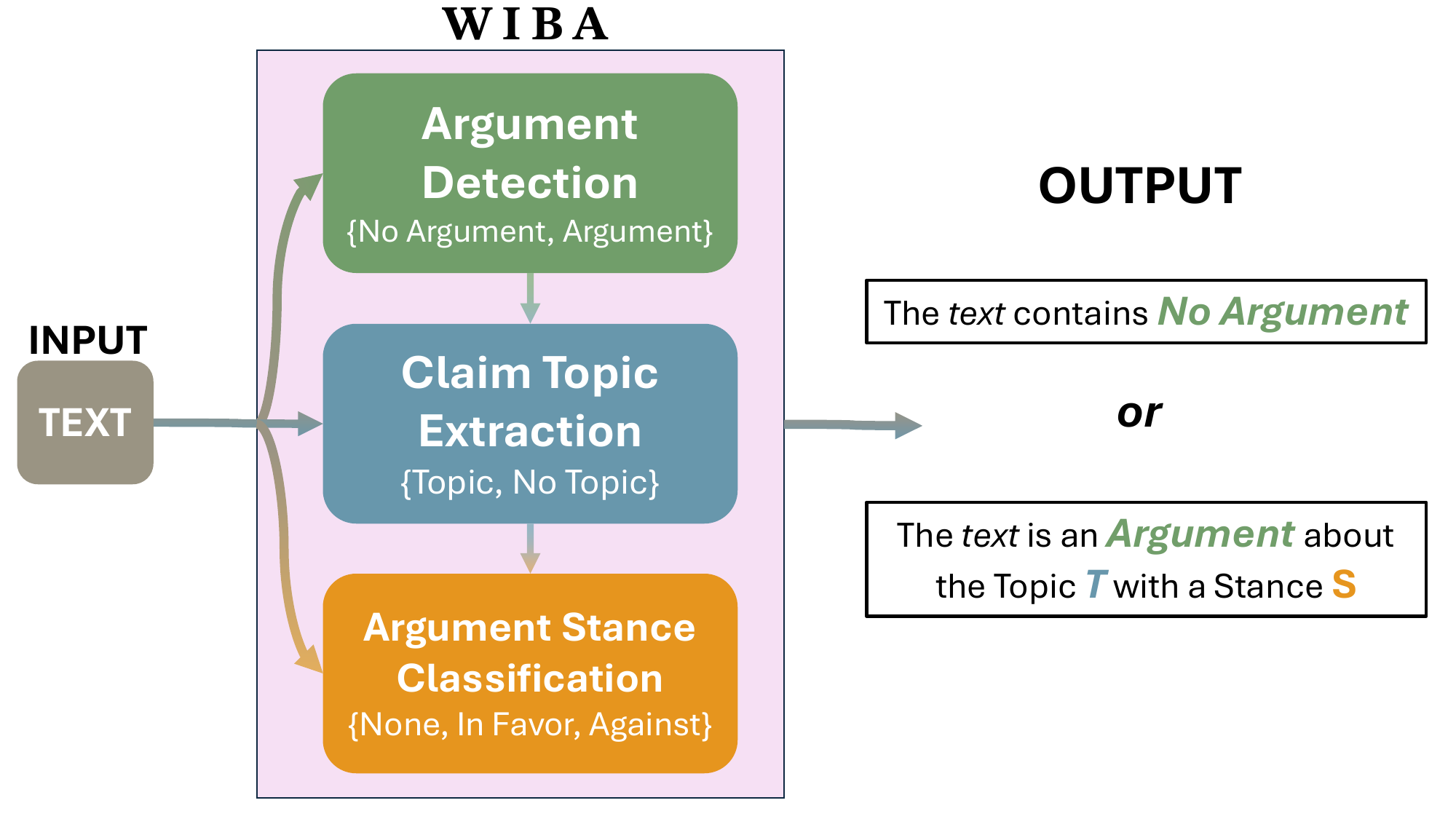}
    \caption{This figure illustrates the proposed task and methods. The process begins with a given text input, where the first step is to determine the presence of an argument. If an argument exists, the subsequent steps involve extracting the topic being argued, and classifying the stance towards that topic. While each method can operate independently, out of order, Argument Stance Classification requires that a specific topic be provided along with the text.}
    \label{fig:pipeline}
\end{figure}

Arguments are fundamental building blocks of effective discourse.
They are critical to decision-making and necessary for forming or altering one's opinion in a variety of communication contexts \cite{Habermas1984}.
This extends from lively conversations among friends and loved ones to heated political debates in formal political forums. As a result, argumentation underwrites democratic legitimacy in many settings \cite{Gutmann1996, Cohen1989}. Despite their linguistic significance, arguments present considerable challenges for detection and quantification through text analytics at scale.

{\bf Opportunity and Challenges:} The emergence of online text-based discourse provides an unprecedented opportunity: there are vast quantities of written opinions, including arguments, that can be mined to better understand what people think and believe, and understanding the reasoning that underwrites their opinions on any topic.
These readily available data are any communication scholar's dream. But there is a catch:
there are several theoretical and technical challenges that  make argument mining difficult. Arguments can be extremely complex in their structure, while at the same time can be extremely sensitive in their semantic appearance, as specific features of different environments may significantly influence how argumentation is expressed. Different environments present different kinds of affordances for arguers \cite{halpernandgibs}. An argument made in a legal procedure will look and read vastly different compared to an argument within a Reddit post. 

\textbf{Problem:} How can we model argumentation in text-mediated discourse? To automatically and accurately identify argumentative characteristics of texts at scale across various digital environments introduces a set of complex challenges to overcome. For example, argumentation in online public forums is often informal and requires a level of contextual understanding in order to determine whether a post meets the formal conditions to be classified as an argument (stated below), and then to understand the content of the argument. In addition, it is challenging to develop a formalization of argumentation that can even be applied to an online public communication that is generalizable, yet specific enough to distinguish the fine line between a non-argument and a low-quality argument. Finally, developing tools that can meet these requirements to accurately understand the necessary argumentative characteristics poses several design challenges.  

\textbf{Previous Work:} 
There has been relatively few efforts that have combined: (a) a comprehensive treatment of arguments, and (b) extensive computational and algorithmic development, especially with open-source tools.
We can group previous works in a few large categories: 
(a) communications, linguistics and philosophy-oriented efforts,
(b) algorithmic and implementation-oriented approaches, but typically with a niche focus,
and (c) Large Language Model (LLM) approaches.
We review related work and discuss how our work differs from them in section~\ref{sec:related}.

{\bf Contribution:} 
 We propose WIBA, a systematic approach to enable the comprehensive  understanding of \textbf{W}hat \textbf{I}s \textbf{B}eing \textbf{A}rgued.
 At a high level, our approach has two distinguishing features.
 First, it is comprehensive as we address all three questions regarding arguments: existence, topic, and stance toward the topic.
 Second, we develop algorithmic solutions that outperform previous techniques, with an ability to handle both informal (e.g., Reddit or Twitter) and formal (e.g., legal or political proceedings) types of arguments. Our approaches leverage LLMs, which we fine tune and prompt-engineer appropriately.

\textit{a. Framework Contributions.}
We develop a comprehensive framework, 
which introduces a theoretical foundation for computational argument mining. The key contributions of our framework include: (a) a proper formalization of arguments, shown to improve task performance when integrated with LLM prompt engineering; and (b) a ``less is more'' novel approach to task fine-tuning, which improves performance, increases consistency across new and unseen datasets while lowering the barrier of entry for future research which no longer requires massive datasets for task training; (c) we demonstrate that WIBA can identify arguments effectively, regardless of the argument type used or whether the syntax is informal or formal. Types of arguments include inductive, deductive, abductive, analogical, and fallacious. Informal and formal syntax refer to the syntactical style of communication, such as those found in casual online forums or formal legal proceedings. Further elaboration on the definitions of argument type and syntactical structure can be found in the WIBA Framework section.

\textit{b. Algorithmic Contributions.}
From an algorithmic point of view, 
we develop three methods.
{\it Method 1:} we present a fine-tuned argument detection language model that takes a text input, and outputs whether the text is an argument or not, with an $F_1$ score between 80\% and 86\%, across three benchmark datasets, a 20 -- 45pp improvement over existing state-of-the-art methods \cite{ukpclassification}. 
{\it Method 2:} We create a language model that has been trained to identify, for any text that meets the formal definition of an argument, the subject of what is being argued. This model is effective at extracting both implicitly and explicitly mentioned topics and aspects, with an efficacy advantage of nearly 40\% over current keyword and topic modeling methods.
{\it Method 3:} We develop a method that takes a topic and text as input and determines, if the text is an argument, the argumentative stance directed towards the topic provided. This method outperforms existing state-of-the-art methods \cite{ukpclassification} by 12 to 20pp with an $F_1$ score ranging between 71\% and 78\%.

\textit{c. Infrastructure Contributions.}
To maximize the practical impact of our research, we developed and launched the WIBA framework as an accessible online platform, depicted in figure~\ref{fig:wibadev} and further discussed in section~\ref{sec:discussion} (https://wiba.dev). This platform enables researchers to upload files for analysis and, with a single click, apply any of the three WIBA methods discussed in this paper. The results are then made available for download.
In addition, to support ongoing research in this field we: (a) open-source our code,
(b) share our fine-tuned models, and
(c) provide our datasets to the community.

\section{WIBA Framework}
{\bf What is an Argument?} An argument is defined as a piece of text or speech that contains a claim asserting something in favor or against a subject, supported by premises. In order for the argument to meet this formal definition, it must contain at least one claim AND at least one premise to support that claim. A claim is a statement that depends on a premise. Premises are statements that provide evidence, reasons, and support for the claim. In this work, we operate under the assumption that arguments are limited to 1-3 sentences, as that is the range of our data. In our training data, a text instance has an average of 27 words and an average of 2 sentences. 

Our methods identify arguments and their contents, but do not make an assessment of the validity or truth of the arguments. Such an assessment is not necessary for our methodological purposes, nor is it normatively necessary; for example, democracies are designed to enable true and false arguments to compete rather than to have some third party determine their validity \cite{Jefferson1823}.

\textbf{Types of Arguments.} Arguments can take many forms depending on their linguistic structure. Based on discourse theory, every argument conforms to one of the following five types: (a) deductive, (b) inductive, (c) abductive, (d) analogical, or (e) fallacious \cite{stanford-phil}. Below are brief definitions and examples for each type.
 
\noindent
\textit{\textbf{Deductive}} arguments require the claim of the argument to necessarily follow from in the premises, in that if one takes the premises to be true, the claim also must be true. For example, ``Glyphosate is a chemical in GMOs and Glyphosate is bad for you, therefore GMOs are bad for you." 

\noindent
\textit{\textbf{Inductive}} arguments require past observations or common knowledge to provide the necessary justification for the claim. For example, ``Every time I recycle my waste, I reduce my carbon footprint, so recycling must help protect the environment." 

\noindent
\textit{\textbf{Abductive}} arguments require that observations made as evidence provide a plausible explanation for the claim. For example,``Given the proliferation of misinformation and social media algorithms, it is plausible to suspect the manipulation of public opinion and polarization by foreign actors." 

\noindent
\textit{\textbf{Analogical}} arguments require that if two claims are considered to be similar, then the truth of the first means the second is true. For example, ``Enforcing dress codes in schools is like forcing everyone to wear uniforms to stop crime - they both make unnecessary rules in the name of control." 

\noindent
\textit{\textbf{Fallacious}} arguments contain both a premise and a claim, but the latter does not follow from the former, such as when the premise is unjust, incorrect, or scientifically invalid. E.g., ``Environmental activists exaggerate the threat of climate change to advance their political agenda, ignoring scientific dissent and manipulating data to justify costly and ineffective policies."

{\bf What is an Argument Topic?} An argument topic refers to the topic implicitly or explicitly central to the claim being made. An argument necessitates an assertion to be made, and this assertion must be made regarding \textit{something}.

{\bf What is an Argument Stance?} 
An argument stance is a multi-class categorization of sentences into one of three classes: No Argument, Argument in Favor, and Argument Against. Stance classification is a difficult problem in computational linguistics due to its subjectively and the necessary reliance on a well-defined topic to which the stance is oriented. 

In this work, we revisit and improve two well known argument mining tasks, {\bf Argument Detection} and {\bf Argument Stance Classification}. We remove a usual prerequisite of Argument Detection, the presence of a topic, and still demonstrate significantly better performance over baseline methods. Additionally, we introduce a novel computational argumentation task, {\bf Claim Topic Extraction}, which extracts the implicit or explicit topic being argued in a text, if an argument is present.

\begin{figure}[t!]
\begin{centering}
    
\begin{adjustbox}{width=0.8\columnwidth}
\begin{tikzpicture}
\node[state, initial,line width=1.5pt,draw=green!100,minimum size=1.2cm] (s) {$\mathcal{S}$};
\node[state, below right of=s,minimum size=1.2cm] (p) {$\mathcal{P^+}$};
\node[state, above right of=s,minimum size=1.2cm] (c) {$\mathcal{C^+}$};
\node[state, right of=c] (p1) {$\mathcal{C^+ P^+}$};
\node[state, right of=p](c1){$\mathcal{P^+C^+}$};
\node[state, right of=s,line width=1.5pt,draw=red!50,minimum size=1.2cm] (na) {$\Tilde{\mathcal{A}}$};
\node[state, right of=na, line width=1.5pt,draw=blue!50,minimum size=1.2cm] (a) {$\mathcal{A}$};
\node[state, right of=p1] (c2) {$\mathcal{C^+P^+}$};
\node[state, right of=c1] (p2) {$\mathcal{P^+C^+}$};

\draw 
(s) edge[line width=1.0pt, "$C$"] (c) 
(s) edge[line width=1.0pt] (na)
(s) edge[line width=1.0pt, "$P$"] (p)
(c) edge[line width=1.0pt] (na)
(p) edge[line width=1.0pt] (na)
(c) edge[loop above,line width=1.0pt, "$C$"] (c)
(p) edge[loop below, line width=1.0pt,"$P$"] (p)
(c) edge[line width=1.0pt,"$P$"] (p1)
(p) edge[line width=1.0pt,"$C$"] (c1)
(p1) edge[loop above, line width=1.0pt,"$P$"] (p1)
(c1) edge[loop below, line width=1.0pt,"$C$"] (c1)
(c1) edge[line width=1.0pt] (a)
(p1) edge[line width=1.0pt] (a)
(c1) edge[above, line width=1.0pt,"$P$"] (p2)
(p2) edge[below, line width=1.0pt] (c1)
(p1) edge[line width=1.0pt,"$C$"] (c2)
(c2) edge[line width=1.0pt] (p1)
(c2) edge[line width=1.0pt] (a)
(p2) edge[line width=1.0pt] (a)
(p2) edge[loop below, line width=1.0pt,"$P$"] (p2)
(c2) edge[loop above, line width=1.0pt,"$C$"] (c2)

;
\end{tikzpicture}
\end{adjustbox}
\caption{Argument Detection Augmented Transition Network (ATN). $\mathcal{S}$ represents the start of the text classification, $\Tilde{\mathcal{A}}$ represents `Not an Argument', $\mathcal{C}$ represents Claim, $\mathcal{P}$ represents Premise, and $\mathcal{A}$ represents Argument. Looped arrows from $\mathcal{C}$ and $\mathcal{P}$ represent recursive calls to Claim and Premise Augmentation Networks. The symbol $^+$ indicate \textit{at least one} I.e., $\mathcal{P^+C^+}$ is the state where there is at least one premise with at least one claim.}
\label{fig:ATN}
\end{centering}

\end{figure}
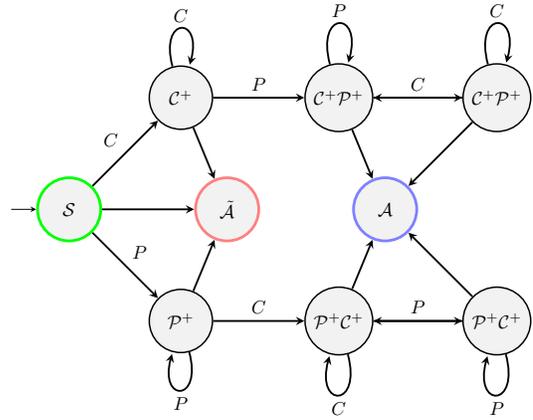

\subsection{Task 1: Argument Detection}

Let $\mathcal{S} = \{x_i, y_i\}$ be a piece of text, where $x_i$ is a sequence of words, and $y_i \in \{0, 1\}$ is a binary label indicating whether $x_i$ constitutes an argument or not. An argument is defined as a sequence of words that semantically contains at least one claim and at least one premise. We define the task of Argument Detection as follows: Given a sequence of words $x_i$, the goal is to determine whether $x_i$ constitutes an argument or not, by checking whether it contains both a claim and a premise.

The task is learned through supervised fine-tuning of LLMs with the goal of minimizing the detection error on unseen argument instances.
We outline in detail the novel formalization of this task and its translation to an Augmented Transition Network in our section \ref{sec:wiba method}. 

\subsection{Task 2: Claim Topic Extraction}
Let $\mathcal{S} = \{x_i, t_i\}$ be a piece of text, where $x_i$ is a sequence of words that may or may not constitute a claim made in an argument, and $t_i$ is the topic being argued. The topic $t_i$ may be explicitly mentioned in the sequence of words $x_i$, or it may be implicitly implied by the argument. We define the task of Claim Topic Extraction from arguments as follows:

Given a sequence of words $x_i$, the goal is to identify the topic $t_i$ in $x_i$, whether the topic is explicitly mentioned or implicitly implied. Formally, we seek to learn a function $f: \mathcal{X} \rightarrow \mathcal{T}$ that maps a sequence of words $x \in \mathcal{X}$ to a topic $t \in \mathcal{T}$, where $\mathcal{X}$ is the set of all possible finite sequences of words, and $\mathcal{T}$ is the set of all possible topics. The function $f$ identifies the topic $t$ of the sequence of words $x$.

Claim Topic Extraction is a novel task within the scope of computational argument mining. In lay terms, Claim Topic Extraction is the ``what'' in the question, ``\textit{What} is being argued?'' Given the formal definition of an argument described earlier, we focus here on the claim rather than on any premises, as the claim is the nucleus of the argument being made. The extraction of what is being argued is crucial in the field of argumentative stance detection. Stance detection not only depends on the presence of a target for attribution, but also is dependent on the contextual accuracy of the target, which in turn controls the accuracy of the model. Claim Topic Extraction distills the argument content and infers the context of the conclusion being made. The motivation for this task is twofold. First, it is often the case that in arguments expressed online, especially in informal exchanges on social media, the topic being argued is often implicit. For example:

\begin{quote}
\textit{``no one should be restricted to only one partner when multiple people may be equally right for them in different ways"}
\end{quote}

\noindent In this example, there is no obvious explicit keyword or topic that is expressed. However, when a human looks at this sentence it is obvious that the topic being argued is monogamy/polygamy. Second, even in the cases where the topic being argued is explicitly mentioned in the claim, there may be more than one potential topics present. For example, 
\begin{quote}
\textit{``The acquisition of nuclear weapons by dictators and despots makes revolution and social change impossible, creating more dire conditions''}
\end{quote}
In this case, there are at least five different explicit topics that potentially could be detected by topic modeling methods: nuclear weapons, dictators, despots, revolution, and social change. Claim Topic Extraction identifies specifically that ``nuclear weapons'' is the topic of the claim, that is, the topic being argued. The other topics are relevant to the evidence or premises of the argument. This distinction is important when considering argument mining as a practical application, in which claim topics may be implicit. Contextual understanding and language generation are crucial for identifying argument topic, making LLMs the optimal architectural choice.

\subsection{Task 3: Argument Stance Classification} 
Let $\mathcal{S} = \{x_i, t_i, s_i\}$ be a piece of text, where $x_i$ is a sequence of words representing a text, $t_i$ is the topic associated with the text, and $s_i \in \{0, 1, 2\}$ is the stance label, where $s_i = 0$ indicates that the text $x_i$ does not contain an argument (No Argument); $s_i = 1$ indicates that the text $x_i$ contains an argument in favor of the topic $t_i$ (Argument in Favor); $s_i = 2$ indicates that the text $x_i$ contains an argument against the topic $t_i$ (Argument Against). We define the task of Argument Stance Classification as follows: Given a sequence of words $x_i$ and a topic $t_i$, the goal is to classify the stance $s_i$ of the text $x_i$ towards the topic $t_i$ as one of the three potential argument stances: No Argument, Argument in Favor, or Argument Against.

The task is learned through supervised fine-tuning of LLMs with the goal of minimizing the classification error on unseen argument instances. The model should be able to identify whether the given text constitutes an argument or not, and if it does, determine whether the argument is in favor of or against the provided topic.

\subsection{Large Language Models}
The following section outlines the basic details of the five Large Language Models we use to evaluate WIBA.\\
\noindent
\textbf{BART}, first introduced in 2020, is a standard Transformer-based neural network \cite{Transformer} which, despite its simplicity, can be seen as generalizing BERT (due to the bidirectional encoder), GPT (with the left-to-right decoder), and other recent pretraining schemes \cite{BART}.

\noindent
\textbf{LLaMa-2 7B/LLaMa-3 8B} are pretrained LLMs released by Meta with 7 billion and 8 billion parameters, respectively\cite{LLaMa2}, \cite{llama3modelcard}. LLaMa-2 is an updated version of LLaMa-1, with its pretraining data corpus being increased by 40\%, doubled context length and grouped query attention. LLaMa-3 8B builds on this by incorporating an additional billion parameters and extra training data, achieving a 20-point improvement over LLaMa-2 7B on the MMLU Benchmark.

\noindent
\textbf{Mistral-7B} is a 7-billion parameter LLM introduced by Mistral.AI, which leverages grouped-query attention and sliding window attention to increase both inference speed and also decoding memory requirements \cite{Mistral}. 

\noindent
\textbf{Yi-6B} is an open-source transformer based LLM that also utilizes LLaMa architecture as the foundational framework. The model does not use LLaMa's weights however, and instead 01.ai created its own proprietary high-quality training datasets and training pipelines.

\section{The WIBA Methodology}
\label{sec:wiba method}
This section is split into three subsections, the first outlines the framework set-up that is used in our Argument Detection and Argument Stance Classification tasks, the second introduces and outlines our proposed novel task, Claim Topic Extraction. And finally, we go over our methodology for Stance Classification. Each of the three proposed methods can be used independently, but can also be utilized sequentially to exploit the dependence: WIBA-Detect $\rightarrow$ WIBA-Extract $\rightarrow$ WIBA-Stance.

\subsection{WIBA-Detect}
\textbf{Data Augmentation.} Recent advances in LLM research propose the value of smaller, higher quality datasets over larger, noisier datasets for specific task fine-tuning \cite{zhou2023lima}. Since modern LLMs have been pretrained on such a large knowledge base, there is no need for the scale and resources of large datasets for task alignment. We adopt this `less is more' approach in our methodology, and train our Language Models on a dataset of only $\sim$1000 instances, for 2 epochs, to prevent overfitting. The LLaMa models alone experience additional benefit from more training for this specific task, and therefore we train these two models for 3 epochs. 

\textbf{Argument Formalization.} An advantage of LLMs is the ability to incorporate prompting instructions for guiding the agent into solving a specific task more effectively. We propose a novel formalization of Argument Detection for the advancement of LLM task prompt engineering. This formalization relies on the structure of a type of nondeterministic finite automaton, Augmented Transition Networks (ATN). This ATN is then translated into a pseudo-language for direct reasoning prompt engineering for both Argument Detection and Argument Stance Classification tasks. The foundation of the ATN is as follows:

\noindent $S$: The input text.\\
$T$: The set of possible tokens that can be derived from $S$. 
Each token represents a subset of words of arbitrary length, that can be one of four possible states, \{Claim, Premise, Not Claim, Not Premise\}. In other words, the set of tokens derived from $S$ is $T(S) = \{t_1,t_2,t_3,...,t_m\}$, where each token $t_i$ is a concatenation of one or more sequential sub-texts. 

Furthermore, a visualization of the ATN with the possible transition states and edge conditions is shown in Figure \ref{fig:ATN} for greater clarity.

\begin{table}[t]
\begin{tabular}{l@{\hskip 0.4cm}cccc}
\toprule
\multicolumn{1}{l}{} & \multicolumn{4}{c}{Benchmark Corpus} \\
\cmidrule{2-5} 
& 
\multicolumn{1}{c}{UKP$_{\text{HQ}}$} & \multicolumn{1}{c}{GPT$_{\text{HQ}}$} & \multicolumn{1}{c}{DEBATE$_{\text{HQ}}$} & \multicolumn{1}{c}{IBM-ARG$_{\text{HQ}}$} \\
\cmidrule(l){2-2} \cmidrule(l){3-3} \cmidrule(l){4-4} \cmidrule(l){5-5}
  Model & $F_1$ & $F_1$ & $F_1$ & $F_1$\\
\midrule 
\rowcolor[HTML]{64e764} Task 1: Detection &  &  &  & \\
\midrule 
\rowcolor[HTML]{FFBA08}\multicolumn{5}{l}{\textbf{Existing Methods}} \\
\rowcolor[HTML]{FFBA08}
\multicolumn{1}{l}{BERT$_{\text{UKP}}$ (+ topic)} & 58.7 & 41 & 59.7 & ---- \\
\rowcolor[HTML]{FFDAF8} 
\midrule 
\multicolumn{5}{l}{\textbf{WIBA-Detect (ATN)}} \\
\rowcolor[HTML]{FFDAF8}
BART & 23 & 44 & 34 & ---- \\
\rowcolor[HTML]{FFDAF8}
Yi-6B & 78.6 & 68 & 67.7 & ---- \\
\rowcolor[HTML]{FFDAF8}
Mistral-7B & 78.7 & 81.6 & 75 & ---- \\
\rowcolor[HTML]{FFDAF8}
LLaMa-2 7B & \textbf{82.4} & 83.3 & 78 & ---- \\
\rowcolor[HTML]{FFDAF8}
LLama-3 8B & 80.2 & \textbf{85.8} &\textbf{ 79.6} & ----\\

\rowcolor[HTML]{FFDAF8} 
\midrule
\multicolumn{5}{l}{\textbf{WIBA-Detect (No ATN)}} \\
\rowcolor[HTML]{FFDAF8}
BART & 23.2 & 27.3 & 35.5 & ---- \\
\rowcolor[HTML]{FFDAF8}
Yi-6B & \textbf{77} & 64 & 67.3 & ---- \\
\rowcolor[HTML]{FFDAF8}
Mistral-7B & 52.3 & 38.5 & 56.8 & ---- \\
\rowcolor[HTML]{FFDAF8}
LLaMa-2 7B & 73.6 & \textbf{77.9} & 54.1 & ---- \\
\rowcolor[HTML]{FFDAF8}
LLaMa-3 8B & 76.4 & \textbf{81.9} & \textbf{69.4} & ---- \\

\midrule \midrule
\rowcolor[HTML]{64e764} Task 3: Stance &  &  &  & \\
\rowcolor[HTML]{FFBA08} \midrule 
\multicolumn{5}{l}{\textbf{Existing Methods}} \\
\rowcolor[HTML]{FFBA08} \multicolumn{1}{l}{BERT$_{\text{UKP}}$ (+ topic)} & 53.2 & 57.3 & ---- & 58.4 \\
\rowcolor[HTML]{FFDAF8} 
\midrule
\multicolumn{5}{l}{\textbf{WIBA-Stance (ATN)}} \\
\rowcolor[HTML]{FFDAF8}
BART & 13.6 & 18.2 & ---- & 20.3 \\
\rowcolor[HTML]{FFDAF8}
Yi-6B & 68.2 & \textbf{78.3} & ---- & 61.4 \\
\rowcolor[HTML]{FFDAF8}
Mistral-7B & 66.8 & 56.8 & ---- & 48.1 \\
\rowcolor[HTML]{FFDAF8}
LLaMa-2 7B & 68.3 & 77.1 & ---- & \textbf{71.3} \\
\rowcolor[HTML]{FFDAF8}
LLaMa-3 8B & \textbf{71} & 75.7 & ---- & 67.1 \\
\bottomrule
\end{tabular}
\caption{WIBA Evaluation Results \label{t:results} }

\end{table}

\subsection{WIBA-Extract}
For this task we fine-tune a LLaMa-3 8B model for 2 epochs, on our CTE dataset, so that it may learn to generate and identify the core topic of what is being argued. We optimize performance through the use of Chain-of-Thought prompting, which has been demonstrated to be useful for symbolic reasoning tasks \cite{chainofthought}. Moreover, we use indirect reasoning prompting, through the use of contrapositives and contradictions in Chain-of-Thought prompting, which has been shown to enhance the overall accuracy of factual reasoning by 27.33\% \cite{contrapositive}. Due to lack of space, we showcase our complete prompt-engineering framework in our repository. For this task, we opt to re-merge our Low-Rank Adaptation (LoRA) \cite{hu2021lora} weights back into the base model for greater contextual understanding capabilities. The final product is a fine-tuned language model capable of extracting the topic being argued in a text, if one exists. 

\subsection{WIBA-Stance}
We fine-tune each of our language models for the three class task of argument stance classification. Using LoRA Adapter, we only tune the linear layers, reducing the number of trainable parameters from $\sim$7 billion to $\sim$20 million. We opt to not re-merge the LoRA weights back into the base model, which has no influence on the performance of the model. This memory-efficient approach enables WIBA-Stance to be used in memory-constrained environments. The model is trained on an input that must contain a topic, and a text in order to anchor a corresponding argumentative stance towards the topic, if there is one. The models are trained in similar fashion to WIBA-Detect, all for 2 epochs, except LLaMa-3 8B which experiences performance improvement until 4 epochs. 

During our research, we observe a much higher order of different types of non-arguments, which initially our model struggles to identify, leading to many false positives. To overcome this issue and improve our $F_1$ performance, we append only to our validation data an additional 150 non-arguments from our IBM-ARG$_{\text{HQ}}$ validation set. 

\textbf{Argument Stance Formalization}. After thorough experimentation of possible interpretations of a stance version of our Argument Transition Network, we find that the same ATN used for Detection performs the best for Stance as well. We therefore rely on the ATN logic in Figure \ref{fig:ATN} for WIBA-Stance.

\section{Evaluation}
Our initial evaluation of WIBA against benchmark datasets returned many false-positives and false-negatives. But when examining the labels we found many instances where sentences in the benchmark data were coded as arguments that do not match our definition of what an argument is, as well as many argument sentences that were coded as not arguments. To resolve this issue, we conduct a blind human labeling of 500 random and even class distributed instances from the benchmark datasets. Each sample is blindly hand re-labeled according to our formal definition of an argument (we provide the Python code to do this blind classification in our replication materials). To validate benchmark methods, we evaluate UKP's BERT based tool on these new subsets of the data to get an accurate understanding of the improvement of the existing tools to WIBA. We next describe our datasets. Then we summarize the results of our evaluation, shown in Table \ref{t:results}.

\subsection{Datasets}
For validation, we use three industry standard benchmark datasets and one we generate synthetically using Chat-GPT.

\textbf{UKP}$_\text{HQ}$: This corpus represents a subset of the full UKP Sentential Argument Mining Corpus by \cite{ukp-corpus}. The original dataset contains 27,520 sentences spanning eight controversial topics. Each sentence is assigned a label of (No Argument, Argument In Favor, Argument Against) a topic. Upon manual investigation of this full dataset, we identified an inconsistency of the sentences with their labels, according to both \cite{ukp-corpus} and our definition of what an argument is. We filter and select high quality examples for our train and validation data for a total of 333 instances across the three classes and an even distribution of the topics within those classes. For the test dataset we randomly selected 504 instances evenly selected across topics and labels, and in a blind-review hand-coded the labels carefully according to the WIBA formalization. 

\textbf{DEBATE}$_{\text{HQ}}$: This IBM Debater dataset consists of 700 annotated argumentative sentences from recorded debates \cite{shnarch2020unsupervised} over 20 different topics. Due to the dataset only containing 260 instances labeled as arguments, we select a subset of 250 non-arguments, and 250 arguments for a total of 500 instances to manually label in our blind-review. Since this dataset only contains a binary argument label, we do not evaluate argument stance classification against this dataset.

\textbf{IBM-ARG}$_{\text{HQ}}$: This dataset originally contained 5.3k arguments with an associate argument quality rank, a continuous value between [0,1] and a (pro, con) stance towards a topic \cite{toledo2019automatic}. We blind review a randomly selected sample of 500 instances with an evenly selected distribution of rank, and accessed each instance carefully to see if the text is an argument or not, towards the topic, and with the appropriate stance. Since this dataset is a multi-class stance dataset, we do not evaluate the argument detection methods against it.

\textbf{GPT}$_{\text{HQ}}$: This dataset consists of 800 short-form examples of synthetically generated arguments and non-arguments, across 400+ topics. The argumentative portion of the dataset contains examples from each of the five different types of reasoning used in argumentation: Inductive, Abductive, Deductive, Fallacious, and Analogical. Furthermore, for each reasoning type, there is a 50/50 split of informal and formal styles. We define the informal style of argumentation to contain a similar syntactical style to those found in online forums, such as Twitter and Reddit. Formal styles of argumentation are similar to those found in legal and political documents, such as congressional hearings or legal proceedings. For the non-arguments, we define a similar stylistic distinction of informal and formal. We prompt ChatGPT-3.5 with  formal definitions for each type of argument \cite{stanford-phil}, and ask it to generate complex and diverse `\textit{argument type'} arguments across a variety of controversial issues. We specify lengths of 1-3 sentences, and either ask to emulate the style of political or legal texts such as congressional hearings, or those found on Reddit and Twitter. The full prompt design is provided in our repository.

\textbf{CTE}$_{\text{Train}}$: This dataset is designed for the task of claim topic extraction, created by combining the instances manually selected from the \textbf{IBM-ARG} dataset and the \textbf{UKP} dataset. Included in the train datasets are a total of 900 arguments with both implicit and explicit claim topics and 150 `No Topic' (No Argument). From the UKP dataset we collect an additional 445 `No Topic' instances. These are split into train, validation, and test splits. 

\subsection{Argument Detection (WIBA-Detect)}
\noindent WIBA-Detect results appear in the top panel of Table \ref{t:results}.

\textbf{The implementation of the WIBA ATN Formalization has a significant positive impact on the $F_1$ Score}. 
By applying WIBAs ATN prompt engineering to assess each LLMs performance and contrasting it with their performance without such logic, a distinct and positive impact is evident, results in enhanced performance across all tested LLMs. BART shows the smallest improvement, only seeing an increase in $F_1$ in the GPT$_{\text{HQ}}$ dataset, likely due to its design limitations regarding system instructions. Mistral-7B experiences the most dramatic improvements in $F_1$ across all the datasets, ranging from 12\%--31\%. LLaMA-2 7B also see's a boost in performance ranging from 5\%--18\%, and LLaMA-3 8B from 4\%-10\%. Given these results across three diverse datasets, it is evident that WIBAs formalization significantly bolsters an LLMs capability to perform Argument Detection in a robust and effective manner.

\textbf{WIBA outperforms existing methods.}
Fine-tuning our LLMs on smaller, high-quality data, without the presence of a topic and only the sentence as the input, yielded an increase in performance compared to the BERT-based fine-tuned model released by the UKP Lab \cite{ukpclassification}. Unsurprisingly, BART is simply unable to adapt to the strict and small training environment of WIBA, as it is an infant language model. All the LLMs have roughly similar $F_1$ scores on the UKP$_{\text{HQ}}$ dataset, however the LLaMa based models excel on the GPT$_{\text{HQ}}$ and Debate$_{\text{HQ}}$ dataset indicating a superior ability to distinguish arguments based on their type and level of formality, and proving to be more adept at detecting formal, spoken arguments. WIBA-Detect consistently outperforms UKP methods using BERT, across all the datasets. For UKP$_{\text{HQ}}$ we see improvements of about 24\%; for GPT$_{\text{HQ}}$ we see improvements ranging from 27\% to 45\%; for Debate$_{\text{HQ}}$ we see improvements from 8\% to 20\%.

\textbf{WIBA excels at identifying different types of arguments.}
Investigating the performance on the GPT$_{\text{HQ}}$ allows us to quantify WIBAs ability to identify different types of arguments with different formality levels. WIBA-Detect correctly identifies 100\% of analogical arguments, 98\% of fallacious argument, 92\% of non-arguments, 87\% of inductive arguments, 74\% of deductive arguments, and 55\% of abductive arguments. In total, evaluation upon this synthetic dataset reveals a strong ability to detect informal arguments, with an accuracy of 76\% and formal arguments with an accuracy of 89\%. Of the abductive and deductive arguments misclassified, 88\% of the texts were very short, informal, and narrative-driven, making it difficult for formalization detection. While WIBAs stellar performance in detecting arguably more valuable argument types (analogical, fallacious) accentuates its robustness, evaluation on this synthetic dataset exposes potential gaps in the training data and methodology that can be addressed in future works.  

\subsection{Claim-Topic Extraction (WIBA-Extract)}
We evaluate WIBA-Extract against popular Keyword Extraction and Topic Modeling methods, such as Flair's Named Entity Recognition (NER) methods \cite{akbik2019flair} and RAKE \cite{inbook}, by taking the average cosine-similarity score, where $\vec{e}$ represents the topic embedding, $\cos$ represents the cosine similarity calculation and $N$ is the length of the test dataset. 
\begin{center}
$\text{CTE Score} = \frac{1}{N} \sum_{i=1}^{N} \cos(\vec{e}_{t_i^{gold}}, \vec{e}_{t_i^{pred}})$
\end{center}

We choose the Sentence Transformer mpnet-base-v2 as our embedding model due to the vast diversity of its training data. If a topic is generated when there is no argument, the cosine similarity for that specific instance will automatically be set to $0$.

\textbf{WIBA-Extract excels in short-form text topic identification.}
Evaluating WIBA on \textbf{GPT}$_{\text{HQ}}$, which contains 402 unique and distinct topics and 800 argument/no-argument instances, WIBA-Extract achieves an impressive CTE Score of 64.8\%. Investigating the performance further, of the instances WIBA correctly identified as arguments, it was able to extract the topic with a CTE Score of 74.2\%. Comparing this to Flair's NER model, the model only generated 41/800 topics against the \textbf{GPT}$_{\text{HQ}}$ dataset, displaying an inability to deal with short texts. Flair therefore only achieves a CTE Score of 2\%; however for the topics it did generate, the similarity of those topics with the gold-label were 55.6\%. RAKE generates 733/800 topics, and achieves an overall CTE Score of 40.4\%, indicating a subpar ability to identify quality topics.

\textbf{WIBA-Extract outperforms current na\"{i}ve topic and keyword modeling techniques. }
Evaluating NER against the CTE$_{\text{Test}}$ dataset, we observe only 72 instances having a topic generated out of all 433 instances. Of the topics generated, Flair NER achieved a CTE Score of 20.1\%. RAKE, although it generates a topic for 397/433 instances, only achieves a CTE Score of 29.3\%. WIBA-Extract on the other hand generated 433 topics, one for every instance. Furthermore, our model demonstrates a remarkable overall CTE Score of 76.8\%, over 40\% higher than the current state-of-the-art techniques. For the correctly identified arguments, WIBA-Extract achieves a CTE score of 83.2\%. 

\subsection{Argument Stance Classification (WIBA-Stance)}
\noindent The WIBA-Stance results are in the bottom panel of Table \ref{t:results}.

\textbf{The implementation of the WIBA ATN Formalization has a significant positive impact on the $F_1$ Score}. Just like for the task of argument detection, the presence of a formalization has a significant impact on the LLM's ability to understand how to correctly classify stances for arguments. Mistral-7B experiences the largest impact in performance, with its $F_1$ score improving by 38\% across all datasets. LLaMa-2 7B also saw significant improvements, ranging from 3\% to 13\%.

\textbf{WIBA-Stance outperforms existing methods.} WIBA-Stance demonstrates improvements from UKPs Classification tool ranging from 13\% to 37\% across benchmark datasets. The most significant improvement was over the GPT$_{\text{HQ}}$ dataset, indicating that WIBA-Stance was able to fill the gap in its ability to identify arguments in different forms.

\begin{figure}
    \centering
    \includegraphics[scale=0.22]{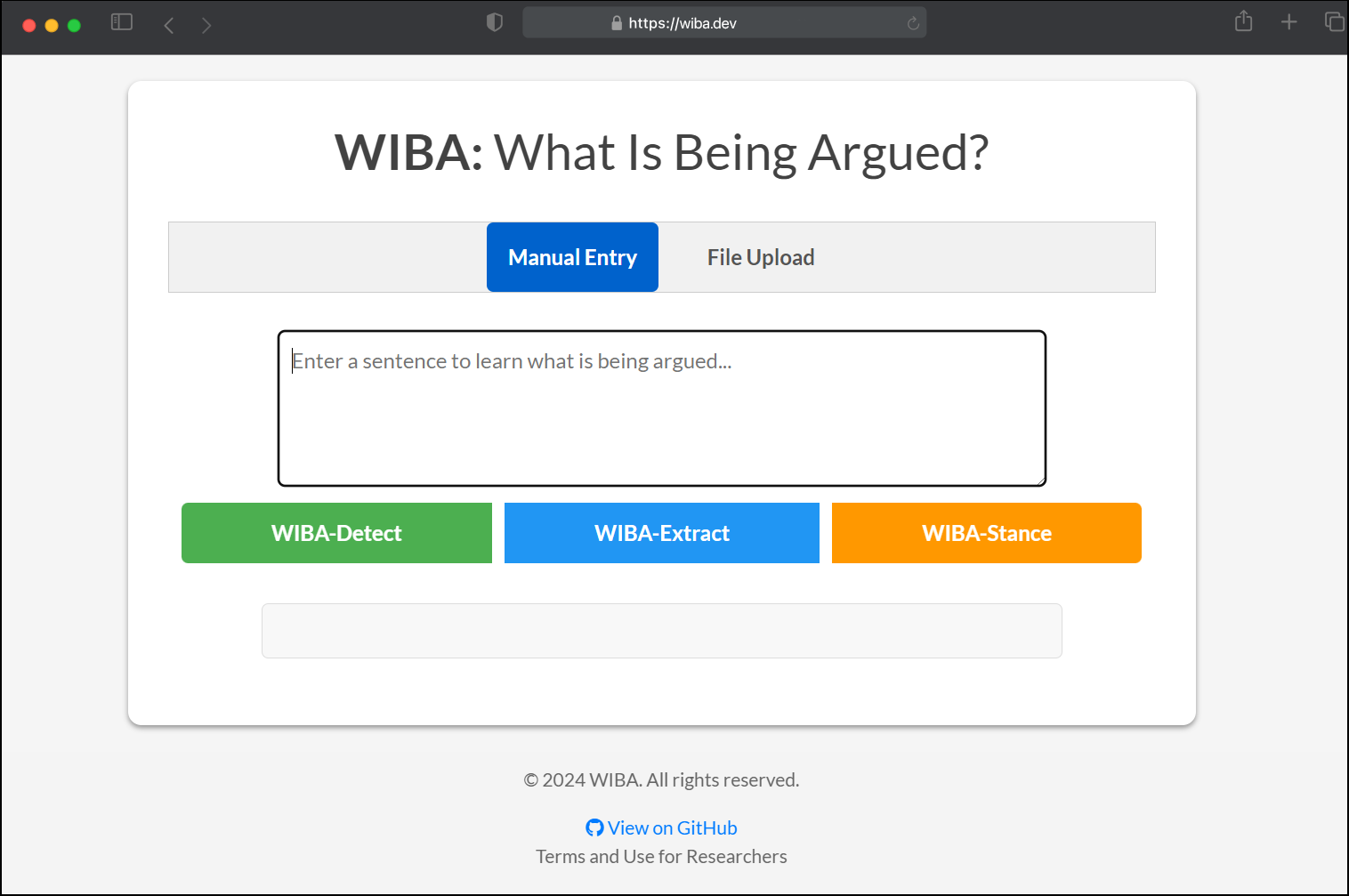}
    \caption{Our platform available at https://wiba.dev}
    \label{fig:wibadev}
\end{figure}

\section{Discussion}
\label{sec:discussion}
{\bf WIBA in action.}
We envision WIBA as a method and tool that will enable the deeper understanding of opinions through argument mining.
Users could include: a) researchers, b) politicians, c) public policy and marketing institutions.
The goal of the user would be to understand in depth what its target group thinks and argues about an issue. We developed a user-friendly platform as seen in \ref{fig:wibadev}, which allows anyone to either manually enter text, or upload large files and perform WIBA tasks.

{\bf What are the limitations of WIBA?}
WIBA is limited by the challenges that any argument mining approach faces. 
These challenges include:
(a) obscure, unconventional and vague language,
(b) multiple intertwined topics and arguments,
(c) detecting sophisticated verbal schemes, mostly sarcasm, and obscure niche pop-culture references.
However, we argue that our innovative three-stage decomposition of the argumentation will improve our ability to clearly identify arguments despite these hurdles compared to existing argument mining methods.

{\bf Can we increase statistical confidence?} 
As with every classification study,
more data and larger ground truth can increase the statistical confidence in our evaluation.
We intend to expand the dataset and the ground truth in the future and continue to share that with the community.

{\bf Will LLMs make WIBA obsolete?}
Our answer to this question is twofold. Firstly, WIBA provides a transparent, structured, and theory-based approach to argument analysis, in contrast to the ``black-box" solution that relying exclusively on an LLM brings. This clear, methodological framework ensures that WIBA retains its relevance and utility, despite the advancements in LLMs. Secondly, WIBA incorporates LLMs as a foundational component, meaning that enhancements in LLM capabilities directly contribute to the improvement of WIBA. For instance, our comparative analysis between LLaMa-3 8B and LLaMa-2 7B reveals that while LLMs continue to evolve, becoming larger and more sophisticated, the performance gains are modest. This suggests that while LLMs are crucial, they do not render WIBA obsolete, but rather enhance its effectiveness.

\section{Related Work}
\label{sec:related}
While there is a great deal of work being conducted on individual components and task of computational argumentation, such as stance detection, claim detection, and counterargument generation, there is little work done on improving the foundational basis upon which all argument mining based tools are built from. Our contributions in this work involve going back to square one, and questioning the validity of the data and definitions a majority of argument mining research has been based on.

\noindent \textbf{Argument Identification } There has been a robust amount of work done on creating models for identifying argumentative components, and for achieving various AM tasks, for example \cite{chen2023exploring} explores the potential advantage of using zero-shot Large Language Models for various computational argumentation tasks. However, these works do not revisit the fundamentals of argument identification, nor investigate the training of these large language models for these tasks. \cite{guo2023aqe} proposes an all-in-one framework for extracting argumentative components i.e., claims, evidence, evidence types, and stances. \cite{levy2014context} proposes a task of context-dependent claim detection. \cite{cheng2021argument} introduces a task for extracting arguments from two passages and identifying potential argument pairs. \cite{palau2009argumentation} introduced the concept of formalizing the argumentation process, but this technique was not generic and robust enough for the various argumentation environments available in the digital age.

\noindent \textbf{Claim Topic Extraction} For our proposed novel task Claim Topic Extraction, there are various related concepts and works, such as \cite{target_stance} which proposes methods for extracting both the target and stance if neither are provided. This work does not require the text be an argument, which introduces a set of challenges for our task CTE. \cite{zhang2021towards,zhang2022bias} demonstrate and define the use of generative aspects for aspect-based sentiment analysis and NER tasks.   

\section{Conclusion}
The key contribution of our work is WIBA, a comprehensive and systematic approach for developing effective argument mining methods. Our solution consists of three state-of-the-art techniques that identify (a) whether a piece of text contains an argument, (b) the topic being argued, and (c) the stance of the argument towards that topic. Our method is grounded on argument theories, and we introduce systematic formalization of arguments and a computational framework which is represented in the form of an Augmented Transition Network. In addition, we introduce the task of Claim Topic Extraction, which to the best of our knowledge is novel. Another important innovation of our research design is that we evaluate WIBA on various types of arguments written in various styles to ensure that it is widely applicable even to the arguments that are short, of diverse types, and informal as frequently observed in online social media forums.

\vspace{.2em}
\bibliography{ref.bib}

\begin{thebibliography}{10}
\providecommand{\url}[1]{#1}
\csname url@samestyle\endcsname
\providecommand{\newblock}{\relax}
\providecommand{\bibinfo}[2]{#2}
\providecommand{\BIBentrySTDinterwordspacing}{\spaceskip=0pt\relax}
\providecommand{\BIBentryALTinterwordstretchfactor}{4}
\providecommand{\BIBentryALTinterwordspacing}{\spaceskip=\fontdimen2\font plus
\BIBentryALTinterwordstretchfactor\fontdimen3\font minus \fontdimen4\font\relax}
\providecommand{\BIBforeignlanguage}[2]{{%
\expandafter\ifx\csname l@#1\endcsname\relax
\typeout{** WARNING: IEEEtran.bst: No hyphenation pattern has been}%
\typeout{** loaded for the language `#1'. Using the pattern for}%
\typeout{** the default language instead.}%
\else
\language=\csname l@#1\endcsname
\fi
#2}}
\providecommand{\BIBdecl}{\relax}
\BIBdecl

\bibitem{Habermas1984}
J.~Habermas, \emph{The Theory of Communicative Action, Volume One: Reason and the Rationalization of Society}, mccarthy~ed.\hskip 1em plus 0.5em minus 0.4em\relax Beacon Press, 1984.

\bibitem{Gutmann1996}
A.~Gutmann and D.~Thompson, \emph{Democracy and Disagreement: Why Moral Conflict cannot be Avoided in Politics, and What Should be Done about It}.\hskip 1em plus 0.5em minus 0.4em\relax Princeton University Press, 1996.

\bibitem{Cohen1989}
J.~Cohen, \emph{Deliberation and Democratic Legitimacy}.\hskip 1em plus 0.5em minus 0.4em\relax Basil Blackwell, 1989, pp. 17--34.

\bibitem{halpernandgibs}
D.~Halpern and J.~Gibbs, ``Social media as a catalyst for online deliberation? exploring the affordances of facebook and youtube for political expression,'' \emph{Computers in Human Behavior}, vol.~29, p. 1159–1168, 05 2013.

\bibitem{ukpclassification}
\BIBentryALTinterwordspacing
N.~Reimers, B.~Schiller, T.~Beck, J.~Daxenberger, C.~Stab, and I.~Gurevych, ``Classification and clustering of arguments with contextualized word embeddings,'' 2019. [Online]. Available: \url{https://arxiv.org/abs/1906.09821}
\BIBentrySTDinterwordspacing

\bibitem{Jefferson1823}
T.~Jefferson, \emph{Statutes at Large in Virginia [1726]}, w.w. henin~ed., 1823, pp. 84--86.

\bibitem{stanford-phil}
C.~Dutilh~Novaes, ``{Argument and Argumentation},'' in \emph{The {Stanford} Encyclopedia of Philosophy}, fall 2022~ed., E.~N. Zalta and U.~Nodelman, Eds.\hskip 1em plus 0.5em minus 0.4em\relax Metaphysics Research Lab, Stanford University, 2022.

\bibitem{Transformer}
A.~Vaswani, N.~Shazeer, N.~Parmar, J.~Uszkoreit, L.~Jones, A.~N. Gomez, L.~Kaiser, and I.~Polosukhin, ``Attention is all you need,'' 2023.

\bibitem{BART}
M.~Lewis, Y.~Liu, N.~Goyal, M.~Ghazvininejad, A.~Mohamed, O.~Levy, V.~Stoyanov, and L.~Zettlemoyer, ``Bart: Denoising sequence-to-sequence pre-training for natural language generation, translation, and comprehension,'' \emph{arXiv preprint arXiv:1910.13461}, 2019.

\bibitem{LLaMa2}
H.~Touvron and L.~M. et. al., ``Llama 2: Open foundation and fine-tuned chat models,'' 2023.

\bibitem{llama3modelcard}
\BIBentryALTinterwordspacing
AI@Meta, ``Llama 3 model card,'' 2024. [Online]. Available: \url{https://github.com/meta-llama/llama3/blob/main/MODEL_CARD.md}
\BIBentrySTDinterwordspacing

\bibitem{Mistral}
A.~Q. Jiang and A.~S. et. al., ``Mistral 7b,'' 2023.

\bibitem{zhou2023lima}
C.~Zhou, P.~Liu, P.~Xu, S.~Iyer, J.~Sun, Y.~Mao, X.~Ma, A.~Efrat, P.~Yu, L.~Yu, S.~Zhang, G.~Ghosh, M.~Lewis, L.~Zettlemoyer, and O.~Levy, ``Lima: Less is more for alignment,'' 2023.

\bibitem{chainofthought}
J.~Wei, X.~Wang, D.~Schuurmans, M.~Bosma, B.~Ichter, F.~Xia, E.~Chi, Q.~Le, and D.~Zhou, ``Chain-of-thought prompting elicits reasoning in large language models,'' 2023.

\bibitem{contrapositive}
Y.~Zhang, Y.~Sun, Y.~Zhan, D.~Tao, D.~Tao, and C.~Gong, ``Large language models as an indirect reasoner: Contrapositive and contradiction for automated reasoning,'' 2024.

\bibitem{hu2021lora}
E.~J. Hu, Y.~Shen, P.~Wallis, Z.~Allen-Zhu, and Y.~Li, ``Lora: Low-rank adaptation of large language models,'' 2021.

\bibitem{ukp-corpus}
\BIBentryALTinterwordspacing
C.~Stab, T.~Miller, B.~Schiller, P.~Rai, and I.~Gurevych, ``Cross-topic argument mining from heterogeneous sources,'' in \emph{Proceedings of the 2018 Conference on Empirical Methods in Natural Language Processing}.\hskip 1em plus 0.5em minus 0.4em\relax Brussels, Belgium: Association for Computational Linguistics, Oct.-Nov. 2018, pp. 3664--3674. [Online]. Available: \url{https://aclanthology.org/D18-1402}
\BIBentrySTDinterwordspacing

\bibitem{shnarch2020unsupervised}
E.~Shnarch, L.~Choshen, G.~Moshkowich, N.~Slonim, and R.~Aharonov, ``Unsupervised expressive rules provide explainability and assist human experts grasping new domains,'' \emph{arXiv preprint arXiv:2010.09459}, 2020.

\bibitem{toledo2019automatic}
A.~Toledo, S.~Gretz, E.~Cohen-Karlik, R.~Friedman, E.~Venezian, D.~Lahav, M.~Jacovi, R.~Aharonov, and N.~Slonim, ``Automatic argument quality assessment--new datasets and methods,'' \emph{arXiv preprint arXiv:1909.01007}, 2019.

\bibitem{akbik2019flair}
A.~Akbik, T.~Bergmann, D.~Blythe, K.~Rasul, S.~Schweter, and R.~Vollgraf, ``{FLAIR}: An easy-to-use framework for state-of-the-art {NLP},'' in \emph{{NAACL}, 2019 Annual Conference of the North American Chapter of the Association for Computational Linguistics (Demonstrations)}, pp. 54--59.

\bibitem{inbook}
S.~Rose, D.~Engel, N.~Cramer, and W.~Cowley, \emph{Automatic Keyword Extraction from Individual Documents}, 03 2010, pp. 1 -- 20.

\bibitem{chen2023exploring}
G.~Chen, L.~Cheng, L.~A. Tuan, and L.~Bing, ``Exploring the potential of large language models in computational argumentation,'' \emph{arXiv preprint arXiv:2311.09022}, 2023.

\bibitem{guo2023aqe}
J.~Guo, L.~Cheng, W.~Zhang, S.~Kok, X.~Li, and L.~Bing, ``Aqe: Argument quadruplet extraction via a quad-tagging augmented generative approach,'' \emph{arXiv preprint arXiv:2305.19902}, 2023.

\bibitem{levy2014context}
R.~Levy, Y.~Bilu, D.~Hershcovich, E.~Aharoni, and N.~Slonim, ``Context dependent claim detection,'' in \emph{Proceedings of COLING 2014, the 25th International Conference on Computational Linguistics: Technical Papers}, 2014, pp. 1489--1500.

\bibitem{cheng2021argument}
L.~Cheng, T.~Wu, L.~Bing, and L.~Si, ``Argument pair extraction via attention-guided multi-layer multi-cross encoding,'' in \emph{Proceedings of the 59th Annual Meeting of the Association for Computational Linguistics and the 11th International Joint Conference on Natural Language Processing (Volume 1: Long Papers)}, 2021, pp. 6341--6353.

\bibitem{palau2009argumentation}
R.~M. Palau and M.-F. Moens, ``Argumentation mining: the detection, classification and structure of arguments in text,'' in \emph{Proceedings of the 12th international conference on artificial intelligence and law}, 2009, pp. 98--107.

\bibitem{target_stance}
Y.~Li, K.~Garg, and C.~Caragea, ``A new direction in stance detection: Target-stance extraction in the wild,'' in \emph{Proceedings of the 61st Annual Meeting of the Association for Computational Linguistics (Volume 1: Long Papers)}, 2023, pp. 10\,071--10\,085.

\bibitem{zhang2021towards}
W.~Zhang, X.~Li, Y.~Deng, L.~Bing, and W.~Lam, ``Towards generative aspect-based sentiment analysis,'' in \emph{Proceedings of the 59th Annual Meeting of the Association for Computational Linguistics and the 11th International Joint Conference on Natural Language Processing (Volume 2: Short Papers)}, 2021, pp. 504--510.

\bibitem{zhang2022bias}
S.~Zhang, Y.~Shen, Z.~Tan, Y.~Wu, and W.~Lu, ``De-bias for generative extraction in unified ner task,'' in \emph{Proceedings of the 60th Annual Meeting of the Association for Computational Linguistics (Volume 1: Long Papers)}, 2022, pp. 808--818.

\end{thebibliography}
\bibliographystyle{IEEEtran}
\end{document}